\documentclass[letterpaper, 10pt, conference]{ieeeconf}  

\IEEEoverridecommandlockouts                              

\usepackage{graphicx}
\usepackage{colortbl}
\usepackage{booktabs}
\usepackage{comment}
\usepackage{times}
\usepackage{epsfig}
\usepackage{lipsum}
\usepackage{amsmath}
\usepackage{amssymb}
\usepackage{arydshln}
\usepackage{color}
\usepackage{caption}
\usepackage{url}
\usepackage{here}
\usepackage{cite}
\usepackage{xspace}
\usepackage{multirow}
\usepackage[ruled,vlined]{algorithm2e}
\usepackage[bookmarks=true, pagebackref]{hyperref}
\let\labelindent\relax
\usepackage{enumitem}

\newcommand{\colornewparts}[0]{\color{black}}
\newcommand{\realsense}[0]{narrow FoV camera\xspace}
\newcommand{\fisheye}[0]{wide FoV camera\xspace}
\newcommand{\spherical}[0]{spherical camera\xspace}
\newcommand\blfootnote[1]{%
  \begingroup
  \renewcommand\thefootnote{}\footnote{#1}%
  \addtocounter{footnote}{-1}%
  \endgroup
}
\newcommand{\MethodName}[0]{ExAug\xspace}
\renewcommand{\baselinestretch}{0.985}

\title{\LARGE \bf
\MethodName: Robot-Conditioned Navigation Policies \\via Geometric Experience Augmentation
}

\author{Noriaki Hirose$^{1,2}$, Dhruv Shah$^{1}$, Ajay Sridhar$^{1}$ and Sergey Levine$^{1}$
\thanks{$^{1}$UC Berkeley, $^{2}$Toyota Motor North America}%
}

\begin{document}
\maketitle
\thispagestyle{empty}
\pagestyle{empty}

\begin{abstract}
Machine learning techniques rely on large and diverse datasets for generalization. Computer vision, natural language processing, and other applications can often reuse public datasets to train many different models. However, due to  differences in physical configurations, it is challenging to leverage public datasets for training robotic control policies on new robot platforms or for new tasks. In this work, we propose a novel framework, \MethodName to augment the experiences of different robot platforms from multiple datasets in diverse environments. \MethodName leverages a simple principle: by extracting 3D information in the form of a point cloud, we can create much more complex and structured augmentations, utilizing both generating synthetic images and geometric-aware penalization that would have been suitable in the same situation for a different robot, with different size, turning radius, and camera placement. The trained policy is evaluated on two new robot platforms with three different cameras in indoor and outdoor environments with obstacles. \blfootnote{\url{https://sites.google.com/view/exaug-nav}}
\end{abstract}

\section{Introduction}
Machine learning methods can be used to train effective models for visual perception~\cite{deng2009imagenet,He_2016_CVPR,mildenhall2021nerf}, natural language processing~\cite{wang2019glue,lewis2019bart}, and numerous other applications~\cite{sidey2019machine,davenport2019potential}. However, broadly generalizable models typically rely on large and highly diverse datasets, which are usually collected once and then reused repeatedly for many different models and methods. In robotics, this presents a major challenge: every robot might have a different physical configuration, such that end-to-end learning of control policies usually requires specialized data collection for each robotic platform. This calls for developing techniques that can enable learning from experience collected across different robots and sensors.


In this work, we focus in particular on the problem of vision-based navigation, where heterogeneity between robots might include different cameras, sensor placement, sizes, etc., and yet there is structural similarity in the high-level objectives of collision-avoidance and goal-reaching. Training visual navigation policies across experience from multiple robots would require accounting for changes in these parameters, and learning navigation behavior for the target robot parameters --- how do we train such a robot-conditioned policy?
%
Akin to the use of data augmentation to learn invariances for generalization in computer vision and NLP~\cite{shorten2019survey, feng2021survey}, we propose a mechanism for extending the capabilities of learning-based visual navigation pipelines by introducing \emph{experience augmentation}, or \MethodName: a new form of data augmentation to learn navigation policies that can generalize to a variety of different robot parameters,  
such as dynamics or camera intrinsics, and is robust to small variations (due to wear or quality control). 
In contrast to typical augmentation schemes that operate on single data points of images or language, since we are dealing with sequential data, \MethodName performs augmentations at the level of robot trajectories. Our key insight is that we can generate such augmentations for free if we have access to the geometry of the scene: using scene geometry, we can simulate what the robot's observations would appear from a different viewpoint, with a different camera hardware, and even what actions the robot should take if it had different physical properties (e.g., size or turning radius).

\begin{figure}[t]
  \begin{center}
      \includegraphics[width=0.99\hsize]{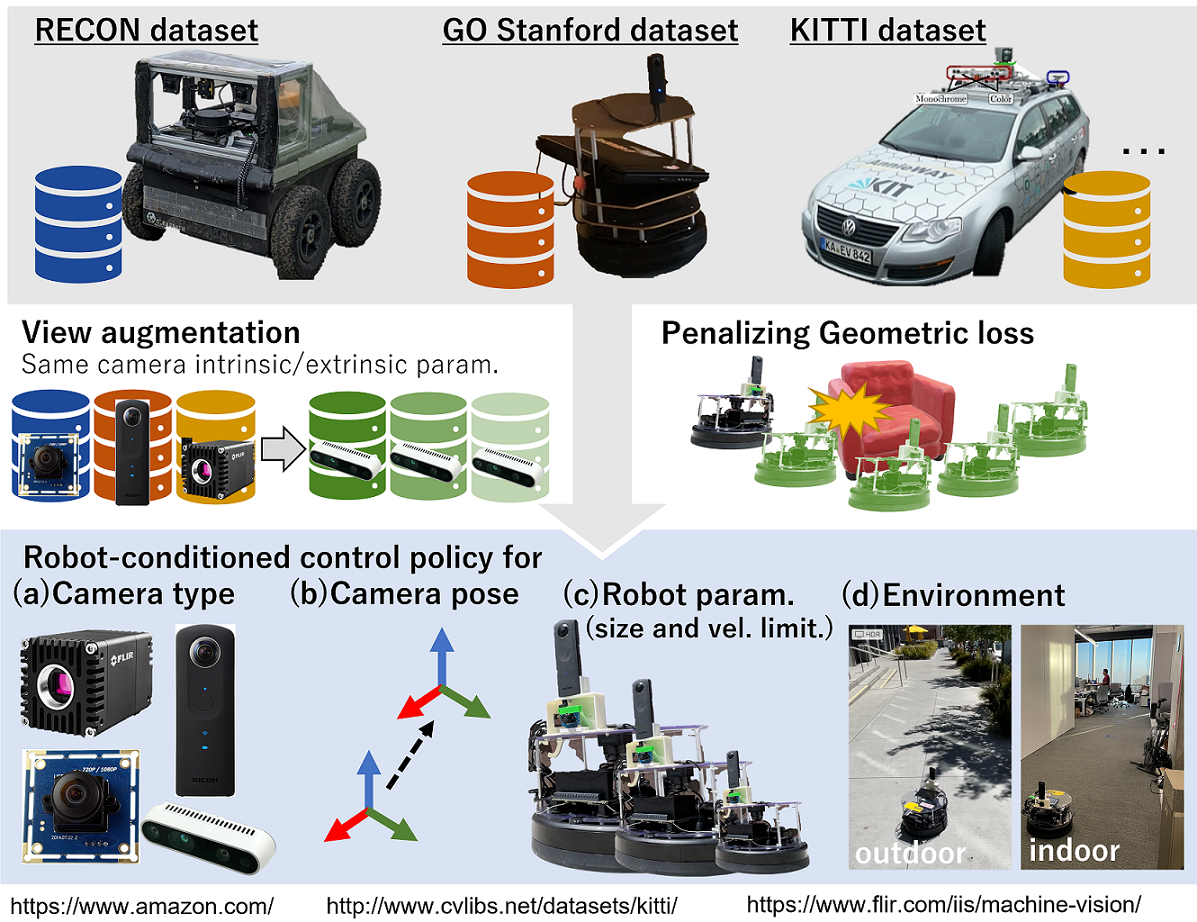}
  \end{center}
      \vspace*{-3mm}
	\caption{\small {\bf Learning navigation via 3D geometric experience augmentation.} Recovering a point cloud allows the system to imagine what the same situation would look like for a different robot, and what that different robot would need to do in that situation.}
  \label{f:pull}
  \vspace*{-8mm}
\end{figure}

\MethodName uses a combination of self-supervised depth estimation to recover scene geometry, novel view synthesis for generating counterfactual robot observations, and training a policy assuming different robot parameters, to augment the robot behavior in the public datasets --- in essence, giving us infinite data from the training environments. We train a shared navigation policy on a combination of trajectories from three distinct datasets representing indoor navigation, off-road driving, and on-road autonomy, on vastly different robot platforms with different sensor stacks --- augmented with \MethodName (Fig.~\ref{f:pull}),
and show that the trained policy can successfully navigate to visual goals from novel viewpoints, in novel environments, as well on novel robot platforms, by leveraging the various affordances depicted by the different datasets.

The primary contribution of this paper is \MethodName, a novel framework to augment the robot experiences in the multiple datasets to train multi-robot navigation policies. We successfully deploy policies trained with \MethodName on two different robots in a variety of indoor and outdoor driving environments, and show that such policies can generalize across a range of robot parameters such as camera intrinsics (e.g. focal lengths, distortions), viewpoints, and dynamics (e.g. robot sizes or turning radii).



\section{Related Work}

Geometric approaches to vision-based navigation have been well-studied in the robotics community for tasks such as visual servoing~\cite{hutchinson1996tutorial,chaumette2006visual,chaumette2007visual}, model predictive control~\cite{li2015vision, sauvee2006image}, and visual SLAM~\cite{mur2017orb, kim2013perception}. However, such geometric methods rely strongly on the geometric layout of the scene, often being too conservative~\cite{kahn2021badgr}, and are susceptible to changes in the map due to dynamic objects.

Learning-based techniques have been shown to alleviate some of these challenges by operating on a topological, and not geometric, representation of the environment~\cite{savinov2018semi, chen2019behavioral, shah2021ving}.
Other methods have also successfully used predictive models~\cite{hirose2019deep, pathak2018zero}, representation learning~\cite{zhu2017target, kahn2018self}, and probabilistic representations to improve navigation over long distances~\cite{hirose2021probabilistic, shah2022viking}. While these approaches have been shown to generalize to environmental modifications, they strongly rely on the deployment trajectories (visual observations, robot dynamics, viewpoint, etc.) to be ``in-distribution'' with respect to their training datasets, which are usually collected on a single robot platform. Our key insight is that we can develop an experience augmentation technique that utilizes data from various robots and augments it to support training policies conditioned on robot parameters that generalize even to new robot configurations.

Data augmentation is a popular mechanism used to artificially increase the ``span'' of a dataset to encourage learning of ``invariances'', hence avoiding over-fitting to a small training dataset~\cite{shorten2019survey, feng2021survey}. A popular use-case of augmentations in robotics is to use a simulated environment~\cite{xia2018gibson,savva2019habitat}, with randomized augmentations to boost data diversity. However, transferring policies from sim-to-real has a number of challenges due to the inability to simulate complex, real-world environments, and dynamic environmental changes~\cite{kadian2020sim2real}. Some recent work has also studied augmenting the training dataset with non-natural images~\cite{Kataoka_2020_ACCV, nakashima2022can, wang2022visual}. \MethodName proposes to use augmentations in a similar spirit to these works, but instead of applying simple image-space transformations, it applies augmentations at the level of entire trajectories by modifying both the observations as well as actions.

The closest concurrent works to \MethodName are Ex-DoF~\cite{tahara2022ex}, which studies synthetic view generation of spherical to augment training data but does not generalization across sensors and other physical robot parameters, and GNM~\cite{shah2022gnm}, which studies direct generalization of simple goal-conditioned policies across different robots simply by training on highly diverse datasets. In contrast, \MethodName studies how geometric understanding can be used to augment \emph{experience}, in the form of counterfactual views and actions, and trains a  policy conditioned on the robot's configuration.

\section{Training Generalizable Policies with \MethodName}
\MethodName uses a geometric framework for data augmentation, where a local point cloud representation of the scene is reconstructed from monocular images, and then used to synthesize novel views to augment the dataset observations. This point cloud is also used to learn novel behavior via our proposed geometry-aware policy objective, providing supervision for the policy conditioned on \emph{counterfactual} robot configurations (e.g., if a larger robot were in the same place, what would it do?). 
These two techniques together allow us to augment the robot experiences in multiple datasets to train the policy with counterfactual images and actions. Goal-conditioned policies trained with \MethodName can thus be deployed on robots with novel parameters, such as new camera viewpoints, robot sizes, turning radii etc. By combining our method with a topological graph~\cite{savinov2018semi, shah2021ving, meng2020scaling}, we obtain a system that performs long-range navigation in diverse environments.

\subsection{Novel View Synthesis via Recovered 3D Structure}
\label{sec:augmenting_geometry}

\begin{figure}[t]
  \vspace{2mm}
  \begin{center}
      \includegraphics[width=0.99\hsize]{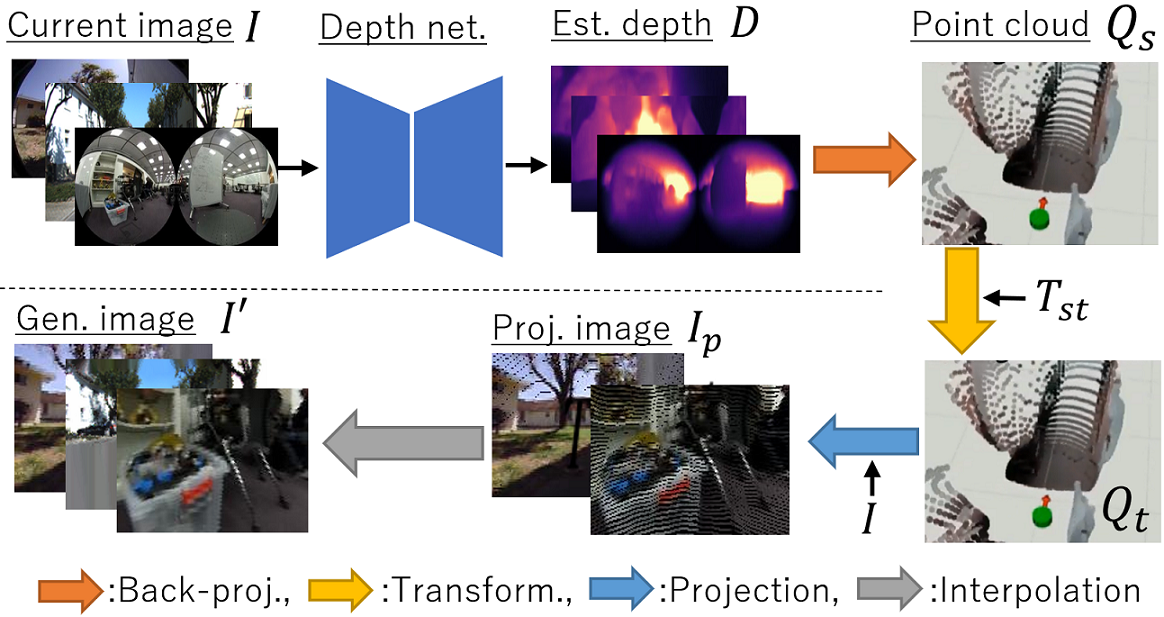}
  \end{center}
    \vspace{-3mm}
	\caption{\small {\bf Overview of view synthesis.} We generate the images by projecting the estimated 3D points on the target image plane with interpolation.}
     \vspace{-6mm}
  \label{f:dataaug}
  
\end{figure}

Figure~\ref{f:dataaug} overviews the synthetic image generation process to train the robot-conditioned policy: given a sequence of images $I$ from a dataset, our goal is to transform them into a sequence $I'$ that matches the parameters of a different camera. For each image, we achieve this by (i) estimating the pixel-wise depth $D$, (ii) back-projecting to a 3D point cloud $Q_s$, and (iii) projecting the point cloud into the image frame of the target camera, resulting in images $I'$. We formulate following process on standard image, camera, and robot coordinates (see. Fig.~\ref{f:vizbot}[e]).

To estimate depth from monocular RGB images, we use the self-supervised method proposed by \cite{zhou2017unsupervised,godard2019digging,hirose2021depth360} and train it on our diverse training dataset from multiple robots. Given access to this depth estimate $D$, we estimate a point cloud $Q_s$ from image $I$ as $Q_s = f_\text{bproj}(D)$, where $f_\text{bproj}$ is the learned camera back-projection function~\cite{hirose2021depth360}. 

To convert this point cloud into a synthetic image $I'$ for a new camera (with different intrinsics and extrinsics),
we first apply a coordinate-frame transformations $T_{st}$ to obtain target-domain point cloud $Q_t$, followed by the target camera projection function $f_\text{proj}$. 
Since the coordinate shift can induce mixed pixels, we use a technique of 3D-warping to project the points $Q_t$ to off-grid coordinates $[i_c, j_c]$, followed by weighted bilinear interpolation to obtain the target-domain pixel values~\cite{huang2020fast,5662013}. Mathematically, this transforms the image $I[i,j] \rightarrow I_p[i_c, j_c]$, which is then \emph{resampled} to obtain the discretized image $I'[i, j]$. To generate synthetic images corresponding to the target camera on the target robot, we only measure $f_\text{proj}$ and $T_{st}$ from the target robot and apply it in the above process.

It should be noted that our approach of view synthesis using self-supervised depth estimation is one of many different ways to synthesize novel views for augmenting the dataset~\cite{zhou2016view, riegler2020free, mildenhall2020nerf, wiles2020synsin}, and the proposed experience augmentation framework is compatible with any of these alternatives.

\subsection{Geometry-Aware Policy Learning}
%

The above process generates synthetic images via perceptual augmentation. We now discuss how the control policy is trained, leveraging the same point cloud representation that we used for augmentation to instead provide supervision suitable for a variety of robot types. The actions in the dataset are specific to the robot that collected each trajectory, and might not be appropriate for a robot with a different size or speed constraint. To train a policy that can generalize effectively over robot configurations, we aim to augment this data and provide supervision to the policy that indicates what a \emph{different} robot would have done in the same situation. To this end, we derive a policy objective that incorporates this \emph{experience augmentation} via a control objective guided by the estimated point cloud.

We design a policy architecture $\pi_\theta$ that predicts velocity commands $\{v_i, \omega_i\}_{i=1 \ldots N_s}$ and traversability estimates $\{t_i\}_{i=1 \ldots N_s}$, corresponding to the inverse probability of collision. We condition this policy on the current and goal observations $\{I_c, I_g\}$, and robot parameters corresponding to size $r_s$ and velocity constraints $v_l$,
i.e., {\colornewparts$\{v_i, \omega_i, t_i \}_{i=1 \ldots N_s} = \pi_\theta(I_c, I_g, r_s, v_l)$, with the policy parameterized by neural network weights $\theta$.}
Note that $I_c$ and $I_g$ are synthetic images generated as per Section~\ref{sec:augmenting_geometry} during training, though at test time the policy uses raw images from the target robot's actual camera. We parameterize the robot as a cylinder with radius $r_s$ and a finite height $h_s$, which is used for estimating collisions with point cloud from original image of $I_c$. We use integrated velocities to estimate future robot positions in the frame of the local point cloud, and define a objective $J$ to train the robot-conditioned policy $\pi_\theta$ encouraging goal-reaching while avoiding collisions:
\begin{equation}
    {\colornewparts \min_{\theta} \,J(\theta) := J_\text{pose}(\theta) + w_{g}J_\text{geo}(\theta) + w_{d}J_\text{diff}(\theta) + w_{t}J_\text{trav}(\theta). }   
    \label{eq:objective}
\end{equation}
%

Here $w_{g}$, $w_{d}$, and $w_{t}$ are weighting factors. To encourage collision-avoidance, we penalize positions that lead to collisions between the cylindrical robot body and the environment points:
\begin{eqnarray}
    J_\text{geo}(\theta) = \frac{1}{L}\sum_{k=1}^{N_s}\sum_{j=1}^{N_V}\sum_{i=1}^{N_U} g[i,j](r_s - d_k[i,j])^2,
    \label{eq:jgeo}
\end{eqnarray}
where $L$ is a normalization factor, $g[i,j]$ is a masked normalization weight that penalizes points inside the robot's body and accounts for the point cloud density, and $d_k[i,j]$ is distance on the horizontal plane between $k$-th predicted waypoint $p_k$ and the back-projection of $I[i,j]$, 
\begin{eqnarray}
    d_{k} = \sqrt{(Q^X_{k})^2 + (Q^Y_{k})^2}.
    \label{eq:dist_3D}
\end{eqnarray}
The point $Q_{k}:=\{Q^X_k, Q^Y_k\}$ on the robot coordinate of $k-$th waypoint is calculated as $Q_{k} = T_{sk} Q_s$, using a coordinate frame transform.


The other components of $J$ correspond to reaching the ground-truth pose $p_{gt}$, predicting the ground-truth traversability $\{t^{gt}_i\}_{i=1\ldots N_s}$, and a smoothness term: 
%
\begin{equation}
\begin{split}
    J_\text{pose}(\theta) &= (p_{gt} - p_{N_s})^2, \\
    J_\text{trav}(\theta) &= \sum_{i=1}^{N_s}(t^{gt}_{i} - t_i)^2, \\
    J_\text{diff}(\theta) &= \sum_{i=1}^{N_s-1}(v_{i+1} - v_i)^2 + (\omega_{i+1} - \omega_i)^2.
\end{split}
    \label{eq:jother}
\end{equation}

To train $\pi_\theta$\, we select pairs of observations $I_c$ and $I_g$ from synthetic images that are less than $N_p$ steps apart. $N_p$ is selected for each dataset based on the robot's speed and dataset's frame rate, to ensure that they are close enough to establish correspondence between the images. 

We assign randomized radii $r_s \in \{r_\text{min}, r_\text{max}\}$m to consider large data diversity in robot size. Besides, we set the robot height $h_s:=(h_\text{max}-h_\text{min})$ to a fixed 0.45 m, with the base $h_\text{min}$ set at 0.2 m to mask out noisy point cloud corresponding to the ground plane. To condition on robot velocity limitation $v_l$, we provide an angular velocity constraint of $\omega_l \in [\omega_\text{min}, \omega_\text{max}]\, \text{rad/s}$. 

By feeding $I_c$, $I_g$, $r_s$ and $v_l$, we can calculate the joint objective $J$ in Eqn.~\ref{eq:objective}. We train $\pi_\theta$ by minimizing $J$ using the Adam optimizer with the learning rate $10^{-3}$. This training process enables us to train \emph{robot-conditioned} policies that can accept different robot parameters.

\section{Implementing \MethodName in a Navigation System}

We now instantiate \MethodName in a navigation system by first implementing the low-level robot-conditioned policy and then integrating it with a navigation system based on topological graphs.

\subsection{Implementation Details: Policy Learning}
For our evaluation systems (see Section~\ref{sec:robots}), we set the limits of variation of the robot radii and the angular velocity as $\{r_\text{min}, r_\text{max}\}$ = $\{$0.0, 1.0$\}$ and $\{\omega_\text{min}, \omega_\text{max}\}$ = $\{$0.5, 1.5$\}$ to adapt to new robots. 
We choose $N_p=8$ steps and weights $\{ w_g, w_d, w_t \}$ to be $\{5e3, 0.025, 0.25\}$ after running hyperparameter sweeps.

\begin{figure}[t]
  \vspace{2mm}
  \begin{center}
      \includegraphics[width=0.99\hsize]{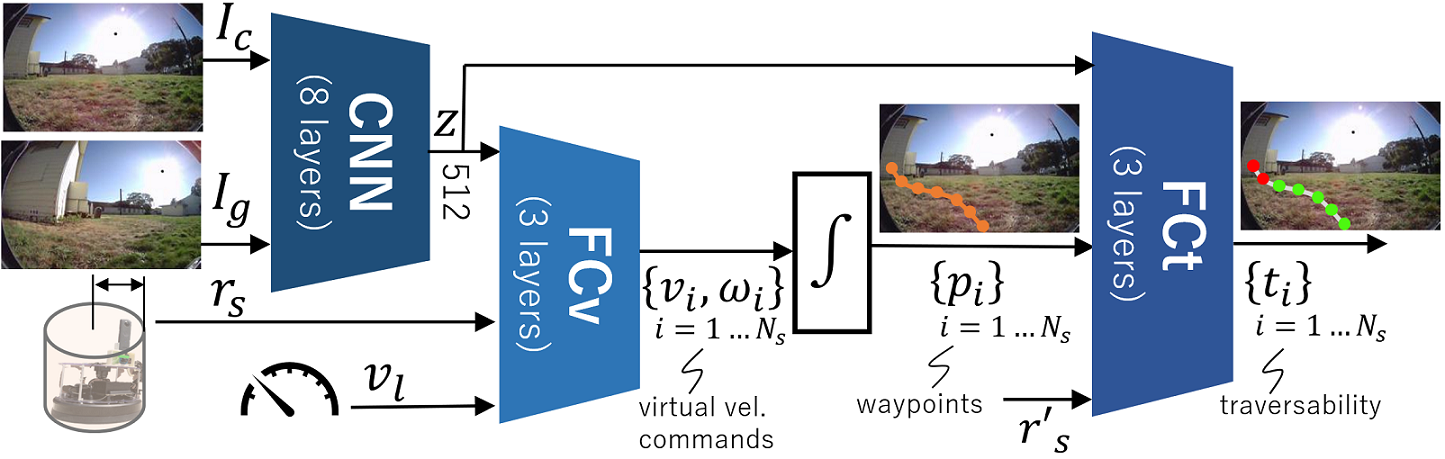}
  \end{center}
      \vspace*{-3mm}
	\caption{\small {\bf Network structure of our control policy $\pi_\theta$.} ``CNN'' extracts image feature $z$ from $I_c$ and $I_g$. ``FCv'' generates the virtual velocity commands conditioned on goal and robot parameters. ``FCt'' estimates traversability at each virtual position.}
  \label{f:network}
  \vspace*{-5mm}
\end{figure}

Figure~\ref{f:network} describes the neural network architecture of $\pi_\theta$. An 8-layer CNN is used to extract the image features $z$ from $I_c$ and $I_g$, with each layer using BatchNorm and ReLu activations. 
The predicted velocity commands $\{v_i, \omega_i\}_{i=1\ldots N_s}$ from 3 fully-connected layers ``FCv'' are conditioned on the robot parameters $\{r_s, v_l\}$ and $z$. A scaled tanh activation is given to limit the output velocities as per the specified constraints $v_l$. 

For estimating traversability $\{t_i\}_{i=1\ldots N_s}$, we integrate the velocities to obtain waypoints predictions and feed them to a set of fully-connected layers ``FCt'' along with the observation embedding $z$ and target robot size $r_s'$, followed by a sigmoid function to limit $t_i \in (0, 1)$. Although $r_s = r_s'$ in training, we found the flexibility of an independent $r_s' \neq r_s$ crucial to the collision-avoidance performance of our system in inference: e.g. setting $r_s = 0.3$ m dictates the conservativeness of the action commands, whereas $r_s'=0.2$ dictates the collision predictions, which can be used as a hard safety constraint for triggering an emergency stop.

\subsection{Navigation system}
%
Since our policy does not allow us to feed the goal image at far position, we construct a mobile robot system that uses the trained policy at the low level, coupled with a topological graph for planning to evaluate in challenging long navigation scenarios~\cite{savinov2018semi}. For the image-goal navigation task, the objective is to navigate to the desired goal image $I_{g_{N_g}}$ by solely relying on egocentric visual observations $I_c$ and searching for a sequence of subgoals $\{ I_{g_i} \}_{i=1\ldots N_g}$ in the topological graph $\mathcal{M}$.

Following \cite{hirose2019deep,hirose2021probabilistic}, the navigation system comprises three ``modules'', tasked with (i) localization, (ii) low-level control, and (iii) safety. For (i), we follow the setup of Hirose et. al.~\cite{hirose2021probabilistic}, where the current observation is localized to the nearest node $i_c$ from the $N_t=5$ adjacent subgoal images, and pass the subgoal image $I_{g(i_c+1)}$ as the \emph{next subgoal} to the policy module. The policy module uses $\pi_\theta$ to obtain velocity commands, which are used in a receding-horizon manner to control the robot, and traversability estimates, which are used by the safety module for emergency stoppage.

\begin{figure}[t]
  \vspace{2mm}
  \begin{center}
      \includegraphics[width=1.0\hsize]{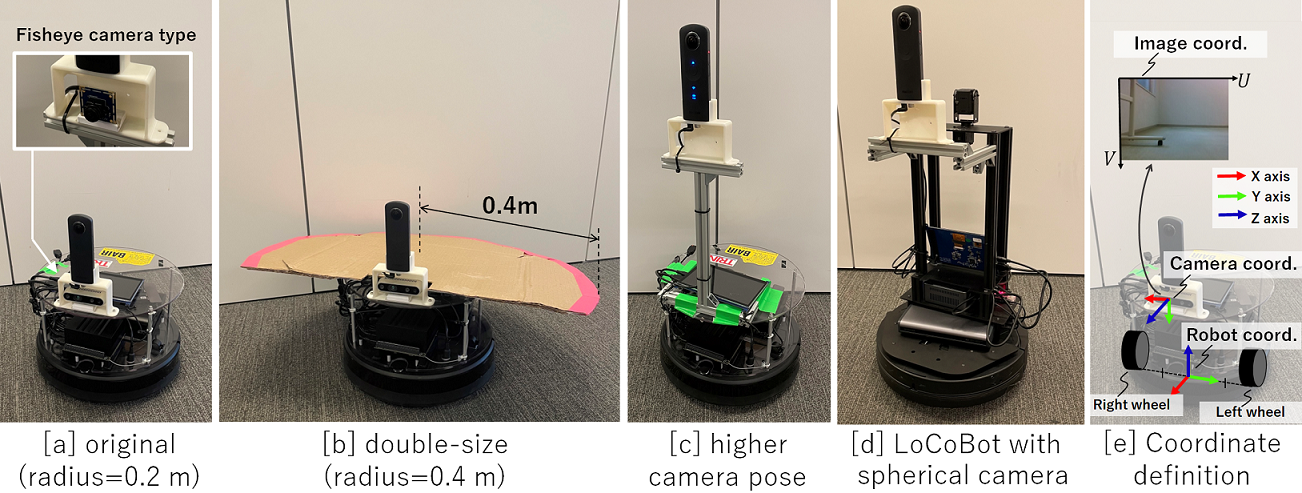}
  \end{center}
      \vspace*{-3mm}
	\caption{\small {\bf The robots in our evaluation.} [a] shows the Vizbot with a \spherical and a \realsense. We can replace the \realsense with the \fisheye. [b] shows a double-size Vizbot with a cardboard body. [c] shows the Vizbot with the \spherical at a higher pose, [d] shows the LoCoBot with a \spherical, and [e] shows the coordinate frame.}
  \label{f:vizbot}
  \vspace*{-5mm}
\end{figure}
\begin{figure*}
  \vspace{2mm}
  \begin{center}
      \includegraphics[width=0.99\hsize]{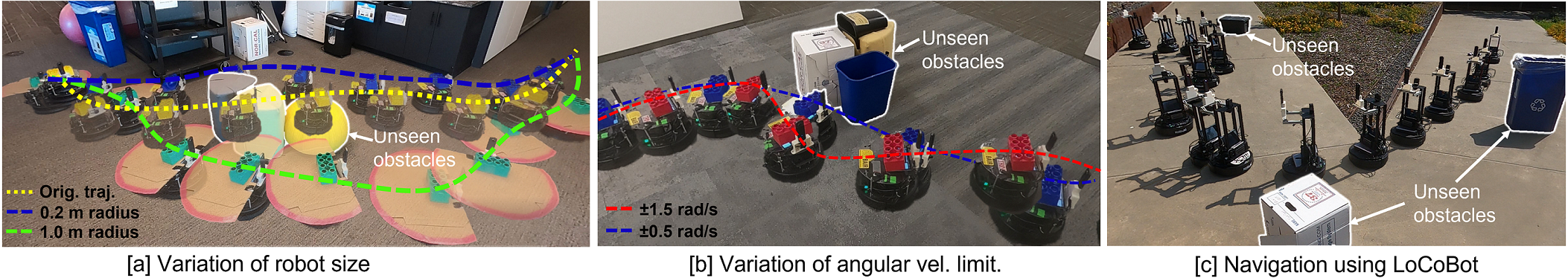}
  \end{center}
    \vspace*{-3mm}
	\caption{\small {\bf Qualitative behavior of \MethodName policies.} ``Orig. traj.'' is the teleoperated trajectory for collecting the graph. [a] Our control policy with $r_s = 1.0$ makes a wider turn to avoid collision, instead of going through the narrow path with $r_s=0.2$. [b] Our control policy with larger angular velocity limits makes a sharper turn to preserve clearance. [c] Our method controls a new robot, the LocoBot, in an outdoor setting.}
	\vspace*{-5mm}
  \label{f:timelapse_all}
\end{figure*}
\section{Multi-Robot Datasets and Setup}
In this section, we describe the datasets used for training the point cloud estimator and our policy, and the robot platforms used for evaluation.
%
\subsection{Training Datasets}

In order to train a generalizable policy that can leverage diverse datasets, we pick three publicly available navigation datasets that vary in their collection platform, visual sensors, and dynamics. This allows us to train policies that can learn shared representations across these widely varying datasets, and generalize to new environments (both indoors and outdoors) and new robots.

\vspace{1mm}
\noindent
\textbf{GO Stanford (GS):} The GS dataset~\cite{hirose2019deep} consists of 10 hours of tele-operated trajectories collected across multiple buildings in a university campus. The data was collected on a TurtleBot2 platform equipped with the Ricoh Theta S \spherical.

\vspace{1mm}
\noindent
\textbf{RECON:} 
The RECON dataset~\cite{shah2021rapid} consists of 30 hours of self-supervised trajectories collected in an off-road environment. This data was collected on a Clearpath Jackal UGV equipped with an ELP fisheye camera.

\vspace{1mm}
\noindent
\textbf{KITTI:} 
KITTI~\cite{Geiger2012CVPR} is a dataset
collected on the roads of Germany, with trajectories recorded from a narrow FoV camera mounted on a driving car. We use a subset of 10 sequences from the KITTI Odometry dataset for training~\cite{kitti_odom}.

\vspace{1mm}
In training, we set the maximum interval $N_p$ between $I_c$ and $I_g$ as 12 for GS and RECON and 2 for KITTI due to the maximum speed of each platform and the frame rate.
\subsection{Implementation Details: View Synthesis}
\label{sec:detail_viewsyn}

Since each dataset contains actions from a single camera, we use the view synthesis process described in Sec.~\ref{sec:augmenting_geometry} to augment the datasets with novel viewpoints. We synthesize views for three target cameras: Intel Realsense (narrow FoV), ELP Fisheye (wide FoV) and Ricoh Theta S (spherical) at a hypothetical camera pose $[0.2, 0.0, 0.3]$ on the robot coordinate to train three policies for each \emph{target} camera. 

We use a self-supervised depth and pose estimation pipeline~\cite{zhou2017unsupervised, monodepth2}. We resize the images for each dataset to a standard size of $416 \times 128 = (N_U \times N_V)$; for data with a spherical camera, we discard the rear-facing camera due to robot body occlusions. Following Hirose et. al.~\cite{hirose2021depth360}, we introduce the learnable camera model to use the dataset without camera parameters and provide supervision for pose estimation during training to resolve scale ambiguities in depth estimates. Please refer to the original paper~\cite{hirose2021depth360} for further implementation details. To generate novel views, we project the 2.5D reconstruction of the scene into a 128$\times$128 image frame for each target camera, obtained by using its camera parameters.

%
\subsection{Evaluation Setup}
\label{sec:robots}
We evaluate policies trained with \MethodName on two \emph{new} robot platforms with a variety of modifications to evaluate different robot sizes, camera hardware, viewpoints etc., as shown in Figure~\ref{f:vizbot}: [a-c] show a custom robot platform, Vizbot~\cite{niwa2022spatio}, which is a new robot without any training data. We deploy the Vizbot in multiple different configurations, such as with a spherical, narrow, or wide FoV camera, artificially boosted robot size (with a cardboard cut-out), and modified camera pose (by mounting the camera on a raised platform). We also evaluate our policies on the LoCoBot platform (Figure~\ref{f:vizbot}[d]), with a different robot base and camera pose.

In addition to closed-loop evaluation, we also perform \emph{offline evaluation} using one hour's worth of navigation trajectories collected with a Vizbot by teleoperating it in both indoor and outdoor environments. Offline evaluation serves as a proxy for closed-loop evaluation during the training process to identify the best hyperparameters. 

%
\begin{figure}[t]
  \begin{center}
      \includegraphics[width=0.99\hsize]{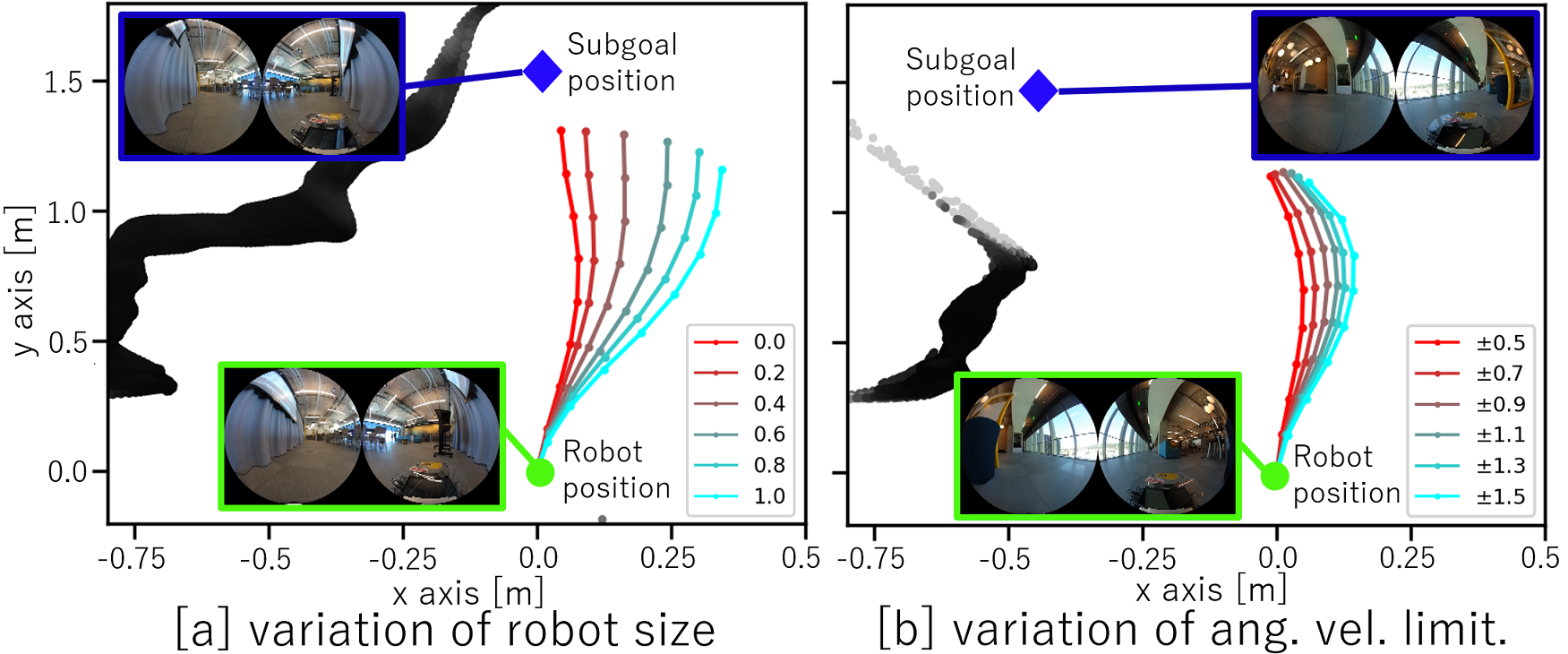}
  \end{center}
      \vspace*{-3mm}
	\caption{\small {\bf Policy outputs for different parameters.} We visualize the generated trajectory from our policy for different robot sizes in [a] and angular velocity constraints in [b].}
  \label{f:control_policy}
  \vspace*{-3mm}  
\end{figure}
\begin{table*}[!t]
  \vspace{2mm}
    \centering
  \caption{{\small {\bf Quantitative comparisons.} In offline evaluation [a], we report the predicted goal-arrival rate (PGA), predicted collision-free rate (PCF), predicted traversability accuracy (PTA), and the full policy objective $J$. In closed-loop evaluation [b], we report the task completion rate (TC), goal arrival rate (GA), and collision-free rate (CF). [a] and [c] use a spherical camera and a RealSense camera for evaluation, respectively. In [b], we evaluate three different cameras. ExAug consistently outperforms the baseline methods in most cases, and in comparing the method with different datasets [c], we find that including all datasets leads to best performance.}}
  \vspace{-1mm}
  \label{tab:qeval}  
  \resizebox{0.68\columnwidth}{!}{
  \begin{tabular}{lcccc}
    \multicolumn{5}{c}{\normalsize [a] {\bf Comparison with baselines in offline data}} \\ \toprule 
    Method & PGA & PCF & PTA & $J^\dagger$ \\ \midrule
    Random & 0.351 & 0.939 & - & 1.483 \\
    Imitation Learning & 0.751 & 0.888 & - & 0.414 \\
    DVMPC & 0.805 & 0.889 & - & 0.358 \\
    \MethodName wo data aug. & 0.801 & 0.929 & \bf{0.897} & 0.367 \\
    \MethodName wo geo. loss & 0.799 & 0.919 & 0.848 & 0.333\\    
    \bf{\MethodName(full)} & \bf{0.833} & \bf{0.945} & 0.858 & \bf{0.319}\\
    \bottomrule
    \multicolumn{5}{r}{$J^\dagger:=J_\text{pose} + w_{g}J_\text{geo} + w_{d}J_\text{diff}$} \\
    \multicolumn{5}{r}{(since the baselines don't predict traversability)}
  \end{tabular}
  }
    \hspace{-2mm}  
  \resizebox{0.65\columnwidth}{!}{
  \begin{tabular}{clcccc}
    \multicolumn{6}{c}{\normalsize [b] {\bf Closed-loop navigation using a real robot.}} \\ \toprule   
    Env. & Method & Cam.type & TC & GA & CF \\ \midrule
    \multirow{4}{*}{\rotatebox{90}{Indoor}} & DVMPC & spherical & 0.678 & 0.389 & 0.611 \\
    & \MethodName & spherical & 0.939 & 0.889 & 0.944 \\
    & & wide FoV & 0.817 & 0.556 & 0.944 \\
    & & narrow FoV & 0.689 & 0.500 & 0.722 \\ \midrule
    \multirow{4}{*}{\rotatebox{90}{Outdoor}} & DVMPC & spherical & 0.528 & 0.278 & 0.889 \\
    & \MethodName & spherical & 0.881 & 0.722 & 0.944 \\
    & & wide FoV & 0.853 & 0.722 & 0.833 \\
    & & narrow FoV & 0.817 & 0.500 & 0.833 \\ \bottomrule  
  \end{tabular}
  }
   \hspace{-2mm}  
    \resizebox{0.7\columnwidth}{!}{
  \begin{tabular}{ccc cccc}
    \multicolumn{7}{c}{\normalsize [c] {\bf Ablation study of dataset in offline data}} \\ \toprule 
    GS & RECON & KITTI & PGA & PCF & PTA & $J$ \\ \midrule
    \checkmark & \checkmark & \checkmark  & \bf{0.674} & \textbf{0.944} & \bf{0.820} & \bf{0.655} \\
    \checkmark & \checkmark & & \bf{0.674} & \textbf{0.950} & \textbf{0.815} & 0.680 \\
    \checkmark & & \checkmark & \textbf{0.669} & \textbf{0.953} & 0.804 & 0.688 \\
    & \checkmark & \checkmark & 0.387 & 0.775 & 0.767 & 1,170\\    
    \checkmark & & & 0.647 & \bf{0.954} & \textbf{0.816} & 0.713 \\
    & \checkmark & & 0.229 & 0.747 & 0.756 & 1.067 \\
    & & \checkmark & 0.206 & 0.751 & 0.759 & 1.177 \\ \bottomrule 
    \\
  \end{tabular}
  }
   \vspace{-4mm}
\end{table*}
\begin{table}[t]
  \caption{{\small {\bf Comparing performances with perturbed viewpoint.} We evaluate the performance of the policies on two different camera height configurations. The results show that \MethodName is robust to viewpoint changes.}}
  \begin{center}
  \vspace{-2mm}
  \resizebox{0.72\columnwidth}{!}{
  \label{tab:rnav2}
  \begin{tabular}{lcccc} \toprule 
    Method & Configuration & TC & GA & CF \\ \midrule
    DVMPC & High (0.6m) & 0.831 & 0.722 & 0.722 \\
     & Low (0.3m) & 0.678 & 0.389 & 0.611 \\
    \MethodName & High (0.6m) & 0.961 & 0.889 & 0.889 \\ 
    & Low (0.3m) & 0.939 & 0.889 & 0.944 \\\bottomrule
  \end{tabular}
  }
  \end{center}
   \vspace{-5mm}
\end{table}
\section{Evaluation}
Our experiments aim to study the following questions:
\begin{enumerate}[label={\bf Q\arabic{*}.}, leftmargin=2.8\parindent]
    \item Can \MethodName augment experience from multiple datasets to train a control policy that generalizes to new robot configurations and robots?
    \item Can \MethodName outperform prior methods, as well as policies trained on single-robot datasets?
\end{enumerate}
\subsection{Comparative Analysis}
We implement \MethodName on a Vizbot and compare it against competitive baselines in a number of challenging indoor and outdoor environments. All our baselines use privileged {\colornewparts omni-directional} observations from a spherical camera,
and we compare the downstream navigation performance against \MethodName with access to different camera observations, including narrow FoV cameras. Despite this, \MethodName consistently outperforms baselines, with a higher success rate and fewer collisions.

\vspace{1mm}
\noindent
\textbf{Random:} 
A control policy that randomly generates velocities between same upper and lower velocity boundaries.

\vspace{1mm}
\noindent
\textbf{Imitation Learning:} 
A control policy to imitate the expert velocity commands by using $\ell_2$ regression. 

\vspace{1mm}
\noindent
\textbf{DVMPC~\cite{hirose2019deep}:} 
A controller based on a velocity-conditioned predictive model that is trained via minimizing the image difference between the predicted images and the goal image. 

\vspace{1mm}
\textbf{Imitation Learning} and \textbf{DVMPC} use raw images from the best single-robot dataset (GS), since sharing heterogeneous data leads to worse performance.
%

In evaluation on an offline dataset of real-world trajectories, Table~\ref{tab:qeval} reports the predicted goal arrival, collision-free rates, traversability accuracy, and test loss values of each method.
From Table~\ref{tab:qeval}[a], we observe that \MethodName consistently achieves higher goal-arrival rates and collision-free rates, as well as a lower loss estimate. We also ablate the augmentation and geometric loss components of \MethodName. 
\MethodName(full) with synthetic images at target camera pose can outperform the ablation of view augmentation with raw images at different camera pose. In addition, our geometric loss performs to avoid collision.

We further compare \MethodName against DVMPC in 12 challenging real-world indoor and outdoor environments for the task of goal-reaching, with 3 trials per environment. Note that DVMPC expects spherical images due to its strong reliance on the geometry of the scene, whereas our method can work with any camera type. In each environment, we collect a topological map by manually teleoperating the robot from start to goal, saving ``image nodes'' at 0.5Hz. We randomly place \emph{novel} obstacles on or between subgoal positions, after {\colornewparts subgoal collection},
in half of the environments.

Table~\ref{tab:qeval}[b] presents the performance of the different variants of \MethodName and DVMPC. Comparing performance with the same camera (spherical), our method vastly outperforms DVMPC: with over a $100\%$ improvement in goal-arrival rates (GA) and $75\%$ reduction in collisions. The performance gap is most significant in challenging environments sharp, dynamic maneuvers and novel obstacles, where DVMPC fails to avoid the obstacle and loses track of the goal. Furthermore, we find that \MethodName continues to perform strongly from cameras with limited FoV, like the RealSense camera, outperforming DVMPC from a spherical camera.

\subsection{Can \MethodName Control Different Robots?}

Next, we qualitatively show the above behaviors
in a closed-loop evaluation with novel robot configurations (Fig.~\ref{f:vizbot}). We systematically evaluate \MethodName on modified versions of a Vizbot with (i) different camera mounting heights, (ii) with different physical sizes, and (iii) with different dynamics constraints. We also show that the same policy can control a LoCoBot, a \emph{new} robot absent in the training datasets.

For evaluating invariance to viewpoint change, we evaluate the two methods on a robot with two different camera height configurations that are 30cm apart. For \MethodName, we train two separate policies, one for each target camera height, and compare against DVMPC. Table~\ref{tab:rnav2} shows that, while DVMPC struggles with goal-reaching and collision avoidance due to the out-of-distribution observations, \MethodName continues to perform strongly and suffers no degradation.

We also perturb the robot size $r_s$ and dynamics constraints on angular velocity $\omega_l$ and find that the trained policies can account for these differences. Fig.~\ref{f:timelapse_all}[a] shows the qualitative behavior of \MethodName under different size parameters: when its radius is increased to 1m, it takes an alternative path around the obstacle because the shortest path would lead to collision with its large footprint, and it successfully reaches the goal. Fig.~\ref{f:control_policy}[a] shows the output of our policy for different values of $r_s$, overlaid on a point cloud cross-section, showing that large $r_s$ leads to increasingly conservative trajectories. Fig.~\ref{f:timelapse_all}[b] shows the qualitative behavior of the robot under different dynamics constraints: when the angular velocity is limited (blue), the robot cannot take sharp maneuvers and instead takes a smoother trajectory to avoid the obstacles. Fig.\ref{f:control_policy}[b] shows the outputs of our policy for different $\omega_l$, suggesting that the policy can account for different constraints.


Lastly,  Fig.~\ref{f:control_policy}[c] shows a timelapse of \MethodName deployed on a LoCoBot. Despite having a higher camera pose and smaller angular velocity range, our method can successfully guide a LoCoBot to the goal while avoiding unseen obstacles in a challenging outdoor environment.
%
%
%
\subsection{Does Training on Multiple Datasets Help?}

A central hypothesis in our work is that, by leveraging datasets from different platforms in different environments and combining them with experience augmentation, we can train generalizable policies that can control robots with different sensors and physical properties. In this section, we evaluate the relative contribution of each portion of our combined dataset to overall performance, so as to ascertain whether more diverse data actually improves performance.

Towards this, we train policies with \MethodName using subsets of the three datasets and evaluate the policies with RealSense on offline real-world trajectories. 
Table~\ref{tab:qeval}[c] shows the 3-dataset policy consistently outperforming other variants, with the lowest loss value (Eqn.~\ref{eq:objective}). We hypothesize that training on larger, more diverse datasets encourages the model to learn a more generalizable representation of the observations, and results in better downstream navigation performance.

\section{Conclusions}

We proposed a framework for training a robot-conditioned policy for vision-based navigation by performing \emph{experience augmentation} on heterogeneous multi-robot datasets. We propose \MethodName, a technique that uses point clouds recovered from monocular images to construct synthetic views and counterfactual action labels to supervise the policy with trajectories that \emph{other} robots would have taken in the same situation. The resulting policy can be conditioned on robot parameters, such as size and velocity constraints, and deployed on new robots and in new environments without additional training. We demonstrate our method on two new robots and in two new environments.

Our method does have a number of limitations. First, we must manually select the robot parameters to condition on. Second, we still require the robots to be structurally similar -- e.g., all robots have forward-facing cameras. An exciting direction for future work would be to utilize the point clouds to also construct synthetic readings for other types of sensors, such as radar, and further extend the range of robots the system can control, potentially with learned latent robot embeddings.

%

%
\bibliographystyle{IEEEtran}
\vskip-\parskip
\begingroup
\footnotesize
\bibliography{egbib}

\begin{thebibliography}{10}
\providecommand{\url}[1]{#1}
\csname url@rmstyle\endcsname
\providecommand{\newblock}{\relax}
\providecommand{\bibinfo}[2]{#2}
\providecommand\BIBentrySTDinterwordspacing{\spaceskip=0pt\relax}
\providecommand\BIBentryALTinterwordstretchfactor{4}
\providecommand\BIBentryALTinterwordspacing{\spaceskip=\fontdimen2\font plus
\BIBentryALTinterwordstretchfactor\fontdimen3\font minus
  \fontdimen4\font\relax}
\providecommand\BIBforeignlanguage[2]{{%
\expandafter\ifx\csname l@#1\endcsname\relax
\typeout{** WARNING: IEEEtran.bst: No hyphenation pattern has been}%
\typeout{** loaded for the language `#1'. Using the pattern for}%
\typeout{** the default language instead.}%
\else
\language=\csname l@#1\endcsname
\fi
#2}}

\bibitem{deng2009imagenet}
J.~Deng, W.~Dong, R.~Socher, L.-J. Li, K.~Li, and L.~Fei-Fei, ``Imagenet: A
  large-scale hierarchical image database,'' in \emph{2009 IEEE conference on
  computer vision and pattern recognition}.\hskip 1em plus 0.5em minus
  0.4em\relax Ieee, 2009, pp. 248--255.

\bibitem{He_2016_CVPR}
K.~He, X.~Zhang, S.~Ren, and J.~Sun, ``Deep residual learning for image
  recognition,'' in \emph{Proceedings of the IEEE Conference on Computer Vision
  and Pattern Recognition (CVPR)}, June 2016.

\bibitem{mildenhall2021nerf}
B.~Mildenhall, P.~P. Srinivasan, M.~Tancik, J.~T. Barron, R.~Ramamoorthi, and
  R.~Ng, ``Nerf: Representing scenes as neural radiance fields for view
  synthesis,'' \emph{Communications of the ACM}, vol.~65, no.~1, pp. 99--106,
  2021.

\bibitem{wang2019glue}
A.~Wang, A.~Singh, J.~Michael, F.~Hill, O.~Levy, and S.~R. Bowman, ``Glue: A
  multi-task benchmark and analysis platform for natural language
  understanding,'' in \emph{7th International Conference on Learning
  Representations, ICLR 2019}, 2019.

\bibitem{lewis2019bart}
M.~Lewis, Y.~Liu, N.~Goyal, M.~Ghazvininejad, A.~Mohamed, O.~Levy, V.~Stoyanov,
  and L.~Zettlemoyer, ``Bart: Denoising sequence-to-sequence pre-training for
  natural language generation, translation, and comprehension,'' \emph{arXiv
  preprint arXiv:1910.13461}, 2019.

\bibitem{sidey2019machine}
J.~A. Sidey-Gibbons and C.~J. Sidey-Gibbons, ``Machine learning in medicine: a
  practical introduction,'' \emph{BMC medical research methodology}, vol.~19,
  no.~1, pp. 1--18, 2019.

\bibitem{davenport2019potential}
T.~Davenport and R.~Kalakota, ``The potential for artificial intelligence in
  healthcare,'' \emph{Future healthcare journal}, vol.~6, no.~2, p.~94, 2019.

\bibitem{shorten2019survey}
\BIBentryALTinterwordspacing
C.~Shorten and T.~M. Khoshgoftaar, ``A survey on image data augmentation for
  deep learning,'' \emph{J. Big Data}, vol.~6, p.~60, 2019. [Online].
  Available: \url{https://doi.org/10.1186/s40537-019-0197-0}
\BIBentrySTDinterwordspacing

\bibitem{feng2021survey}
S.~Y. Feng, V.~Gangal, J.~Wei, S.~Chandar, S.~Vosoughi, T.~Mitamura, and
  E.~Hovy, ``A survey of data augmentation approaches for nlp,'' in
  \emph{Findings of the Association for Computational Linguistics: ACL-IJCNLP
  2021}, 2021, pp. 968--988.

\bibitem{hutchinson1996tutorial}
S.~Hutchinson, G.~D. Hager, and P.~I. Corke, ``A tutorial on visual servo
  control,'' \emph{IEEE transactions on robotics and automation}, vol.~12,
  no.~5, pp. 651--670, 1996.

\bibitem{chaumette2006visual}
F.~Chaumette and S.~Hutchinson, ``Visual servo control. i. basic approaches,''
  \emph{IEEE Robotics \& Automation Magazine}, vol.~13, no.~4, pp. 82--90,
  2006.

\bibitem{chaumette2007visual}
------, ``Visual servo control. ii. advanced approaches [tutorial],''
  \emph{IEEE Robotics \& Automation Magazine}, vol.~14, no.~1, pp. 109--118,
  2007.

\bibitem{li2015vision}
Z.~Li, C.~Yang, C.-Y. Su, J.~Deng, and W.~Zhang, ``Vision-based model
  predictive control for steering of a nonholonomic mobile robot,'' \emph{IEEE
  Transactions on Control Systems Technology}, vol.~24, no.~2, pp. 553--564,
  2015.

\bibitem{sauvee2006image}
M.~Sauv{\'e}e, P.~Poignet, E.~Dombre, and E.~Courtial, ``Image based visual
  servoing through nonlinear model predictive control,'' in \emph{Proceedings
  of the 45th IEEE Conference on Decision and Control}.\hskip 1em plus 0.5em
  minus 0.4em\relax IEEE, 2006, pp. 1776--1781.

\bibitem{mur2017orb}
R.~Mur-Artal and J.~D. Tard{\'o}s, ``Orb-slam2: An open-source slam system for
  monocular, stereo, and rgb-d cameras,'' \emph{IEEE transactions on robotics},
  vol.~33, no.~5, pp. 1255--1262, 2017.

\bibitem{kim2013perception}
A.~Kim and R.~M. Eustice, ``Perception-driven navigation: Active visual slam
  for robotic area coverage,'' in \emph{2013 IEEE International Conference on
  Robotics and Automation}.\hskip 1em plus 0.5em minus 0.4em\relax IEEE, 2013,
  pp. 3196--3203.

\bibitem{kahn2021badgr}
G.~Kahn, P.~Abbeel, and S.~Levine, ``Badgr: An autonomous self-supervised
  learning-based navigation system,'' \emph{IEEE Robotics and Automation
  Letters}, vol.~6, no.~2, pp. 1312--1319, 2021.

\bibitem{savinov2018semi}
N.~Savinov, A.~Dosovitskiy, and V.~Koltun, ``Semi-parametric topological memory
  for navigation,'' in \emph{International Conference on Learning
  Representations}, 2018.

\bibitem{chen2019behavioral}
K.~Chen, J.~P. de~Vicente, G.~Sepulveda, F.~Xia, A.~Soto, M.~V{\'a}zquez, and
  S.~Savarese, ``A behavioral approach to visual navigation with graph
  localization networks,'' \emph{arXiv preprint arXiv:1903.00445}, 2019.

\bibitem{shah2021ving}
D.~Shah, B.~Eysenbach, G.~Kahn, N.~Rhinehart, and S.~Levine, ``Ving: Learning
  open-world navigation with visual goals,'' in \emph{2021 IEEE International
  Conference on Robotics and Automation (ICRA)}.\hskip 1em plus 0.5em minus
  0.4em\relax IEEE, 2021, pp. 13\,215--13\,222.

\bibitem{hirose2019deep}
N.~Hirose, F.~Xia, R.~Mart{\'\i}n-Mart{\'\i}n, A.~Sadeghian, and S.~Savarese,
  ``Deep visual mpc-policy learning for navigation,'' \emph{IEEE Robotics and
  Automation Letters}, vol.~4, no.~4, pp. 3184--3191, 2019.

\bibitem{pathak2018zero}
D.~Pathak, P.~Mahmoudieh, G.~Luo, P.~Agrawal, D.~Chen, Y.~Shentu, E.~Shelhamer,
  J.~Malik, A.~A. Efros, and T.~Darrell, ``Zero-shot visual imitation,'' in
  \emph{Proceedings of the IEEE conference on computer vision and pattern
  recognition workshops}, 2018, pp. 2050--2053.

\bibitem{zhu2017target}
Y.~Zhu, R.~Mottaghi, E.~Kolve, J.~J. Lim, A.~Gupta, L.~Fei-Fei, and A.~Farhadi,
  ``Target-driven visual navigation in indoor scenes using deep reinforcement
  learning,'' in \emph{2017 IEEE international conference on robotics and
  automation (ICRA)}.\hskip 1em plus 0.5em minus 0.4em\relax IEEE, 2017, pp.
  3357--3364.

\bibitem{kahn2018self}
G.~Kahn, A.~Villaflor, B.~Ding, P.~Abbeel, and S.~Levine, ``Self-supervised
  deep reinforcement learning with generalized computation graphs for robot
  navigation,'' in \emph{2018 IEEE International Conference on Robotics and
  Automation (ICRA)}.\hskip 1em plus 0.5em minus 0.4em\relax IEEE, 2018, pp.
  5129--5136.

\bibitem{hirose2021probabilistic}
N.~Hirose, S.~Taguchi, F.~Xia, R.~Mart{\'\i}n-Mart{\'\i}n, K.~Tahara,
  M.~Ishigaki, and S.~Savarese, ``Probabilistic visual navigation with
  bidirectional image prediction,'' in \emph{2021 IEEE/RSJ International
  Conference on Intelligent Robots and Systems (IROS)}.\hskip 1em plus 0.5em
  minus 0.4em\relax IEEE, 2021, pp. 1539--1546.

\bibitem{shah2022viking}
D.~Shah and S.~Levine, ``Viking: Vision-based kilometer-scale navigation with
  geographic hints,'' \emph{arXiv preprint arXiv:2202.11271}, 2022.

\bibitem{xia2018gibson}
F.~Xia, A.~R. Zamir, Z.~He, A.~Sax, J.~Malik, and S.~Savarese, ``Gibson env:
  Real-world perception for embodied agents,'' in \emph{Proceedings of the IEEE
  conference on computer vision and pattern recognition}, 2018, pp. 9068--9079.

\bibitem{savva2019habitat}
M.~Savva, A.~Kadian, O.~Maksymets, Y.~Zhao, E.~Wijmans, B.~Jain, J.~Straub,
  J.~Liu, V.~Koltun, J.~Malik, \emph{et~al.}, ``Habitat: A platform for
  embodied ai research,'' in \emph{Proceedings of the IEEE/CVF International
  Conference on Computer Vision}, 2019, pp. 9339--9347.

\bibitem{kadian2020sim2real}
A.~Kadian, J.~Truong, A.~Gokaslan, A.~Clegg, E.~Wijmans, S.~Lee, M.~Savva,
  S.~Chernova, and D.~Batra, ``{Sim2Real Predictivity: Does Evaluation in
  Simulation Predict Real-World Performance?}'' \emph{IEEE Robotics and
  Automation Letters}, 2020.

\bibitem{Kataoka_2020_ACCV}
H.~Kataoka, K.~Okayasu, A.~Matsumoto, E.~Yamagata, R.~Yamada, N.~Inoue,
  A.~Nakamura, and Y.~Satoh, ``Pre-training without natural images,'' in
  \emph{Proceedings of the Asian Conference on Computer Vision}, 2020.

\bibitem{nakashima2022can}
K.~Nakashima, H.~Kataoka, A.~Matsumoto, K.~Iwata, N.~Inoue, and Y.~Satoh, ``Can
  vision transformers learn without natural images?'' in \emph{Proceedings of
  the AAAI Conference on Artificial Intelligence}, vol.~36, no.~2, 2022, pp.
  1990--1998.

\bibitem{wang2022visual}
Y.~Wang and C.-Y. Ko, ``Visual pre-training for navigation: What can we learn
  from noise?'' \emph{arXiv preprint arXiv:2207.00052}, 2022.

\bibitem{tahara2022ex}
K.~Tahara and N.~Hirose, ``Ex-dof: Expansion of action degree-of-freedom with
  virtual camera rotation for omnidirectional image,'' in \emph{2022
  International Conference on Robotics and Automation (ICRA)}.\hskip 1em plus
  0.5em minus 0.4em\relax IEEE, 2022, pp. 10\,382--10\,389.

\bibitem{shah2022gnm}
\BIBentryALTinterwordspacing
D.~Shah, A.~Sridhar, A.~Bhorkar, N.~Hirose, and S.~Levine, ``{GNM: A General
  Navigation Model to Drive Any Robot},'' in \emph{arXiV}, 2022. [Online].
  Available: \url{https://arxiv.org/abs/2210.03370}
\BIBentrySTDinterwordspacing

\bibitem{meng2020scaling}
X.~Meng, N.~Ratliff, Y.~Xiang, and D.~Fox, ``Scaling local control to
  large-scale topological navigation,'' in \emph{2020 IEEE International
  Conference on Robotics and Automation (ICRA)}.\hskip 1em plus 0.5em minus
  0.4em\relax IEEE, 2020, pp. 672--678.

\bibitem{zhou2017unsupervised}
T.~Zhou, M.~Brown, N.~Snavely, and D.~G. Lowe, ``Unsupervised learning of depth
  and ego-motion from video,'' in \emph{Proceedings of the IEEE conference on
  computer vision and pattern recognition}, 2017, pp. 1851--1858.

\bibitem{godard2019digging}
C.~Godard, O.~Mac~Aodha, M.~Firman, and G.~J. Brostow, ``Digging into
  self-supervised monocular depth estimation,'' in \emph{Proceedings of the
  IEEE/CVF International Conference on Computer Vision}, 2019, pp. 3828--3838.

\bibitem{hirose2021depth360}
N.~Hirose and K.~Tahara, ``Depth360: Monocular depth estimation using learnable
  axisymmetric camera model for spherical camera image,'' \emph{arXiv preprint
  arXiv:2110.10415}, 2021.

\bibitem{huang2020fast}
H.-Y. Huang and S.-Y. Huang, ``Fast hole filling for view synthesis in free
  viewpoint video,'' \emph{Electronics}, vol.~9, no.~6, p. 906, 2020.

\bibitem{5662013}
I.~Daribo and B.~Pesquet-Popescu, ``Depth-aided image inpainting for novel view
  synthesis,'' in \emph{2010 IEEE International Workshop on Multimedia Signal
  Processing}, 2010, pp. 167--170.

\bibitem{zhou2016view}
T.~Zhou, S.~Tulsiani, W.~Sun, J.~Malik, and A.~A. Efros, ``View synthesis by
  appearance flow,'' in \emph{European conference on computer vision}.\hskip
  1em plus 0.5em minus 0.4em\relax Springer, 2016, pp. 286--301.

\bibitem{riegler2020free}
G.~Riegler and V.~Koltun, ``Free view synthesis,'' in \emph{European Conference
  on Computer Vision}.\hskip 1em plus 0.5em minus 0.4em\relax Springer, 2020,
  pp. 623--640.

\bibitem{mildenhall2020nerf}
B.~Mildenhall, P.~Srinivasan, M.~Tancik, J.~Barron, R.~Ramamoorthi, and R.~Ng,
  ``Nerf: Representing scenes as neural radiance fields for view synthesis,''
  in \emph{European conference on computer vision}, 2020.

\bibitem{wiles2020synsin}
O.~Wiles, G.~Gkioxari, R.~Szeliski, and J.~Johnson, ``Synsin: End-to-end view
  synthesis from a single image,'' in \emph{Proceedings of the IEEE/CVF
  Conference on Computer Vision and Pattern Recognition}, 2020, pp. 7467--7477.

\bibitem{shah2021rapid}
D.~Shah, B.~Eysenbach, N.~Rhinehart, and S.~Levine, ``Rapid exploration for
  open-world navigation with latent goal models,'' \emph{arXiv preprint
  arXiv:2104.05859}, 2021.

\bibitem{Geiger2012CVPR}
A.~Geiger, P.~Lenz, and R.~Urtasun, ``Are we ready for autonomous driving? the
  kitti vision benchmark suite,'' in \emph{2012 IEEE conference on computer
  vision and pattern recognition}.\hskip 1em plus 0.5em minus 0.4em\relax IEEE,
  2012, pp. 3354--3361.

\bibitem{kitti_odom}
``Kitti visual odometry benchmark 2012,''
  \url{https://www.cvlibs.net/datasets/kitti/eval_odometry.php)}.

\bibitem{monodepth2}
C.~Godard, O.~{Mac Aodha}, M.~Firman, and G.~J. Brostow, ``Digging into
  self-supervised monocular depth prediction,'' \emph{The International
  Conference on Computer Vision (ICCV)}, October 2019.

\bibitem{niwa2022spatio}
T.~Niwa, S.~Taguchi, and N.~Hirose, ``Spatio-temporal graph localization
  networks for image-based navigation,'' \emph{arXiv preprint
  arXiv:2204.13237}, 2022.

\end{thebibliography}
\endgroup

\section*{APPENDIX}
\subsection{Overview of our method}
\begin{figure}[h]
  \begin{center}
      \includegraphics[width=0.99\hsize]{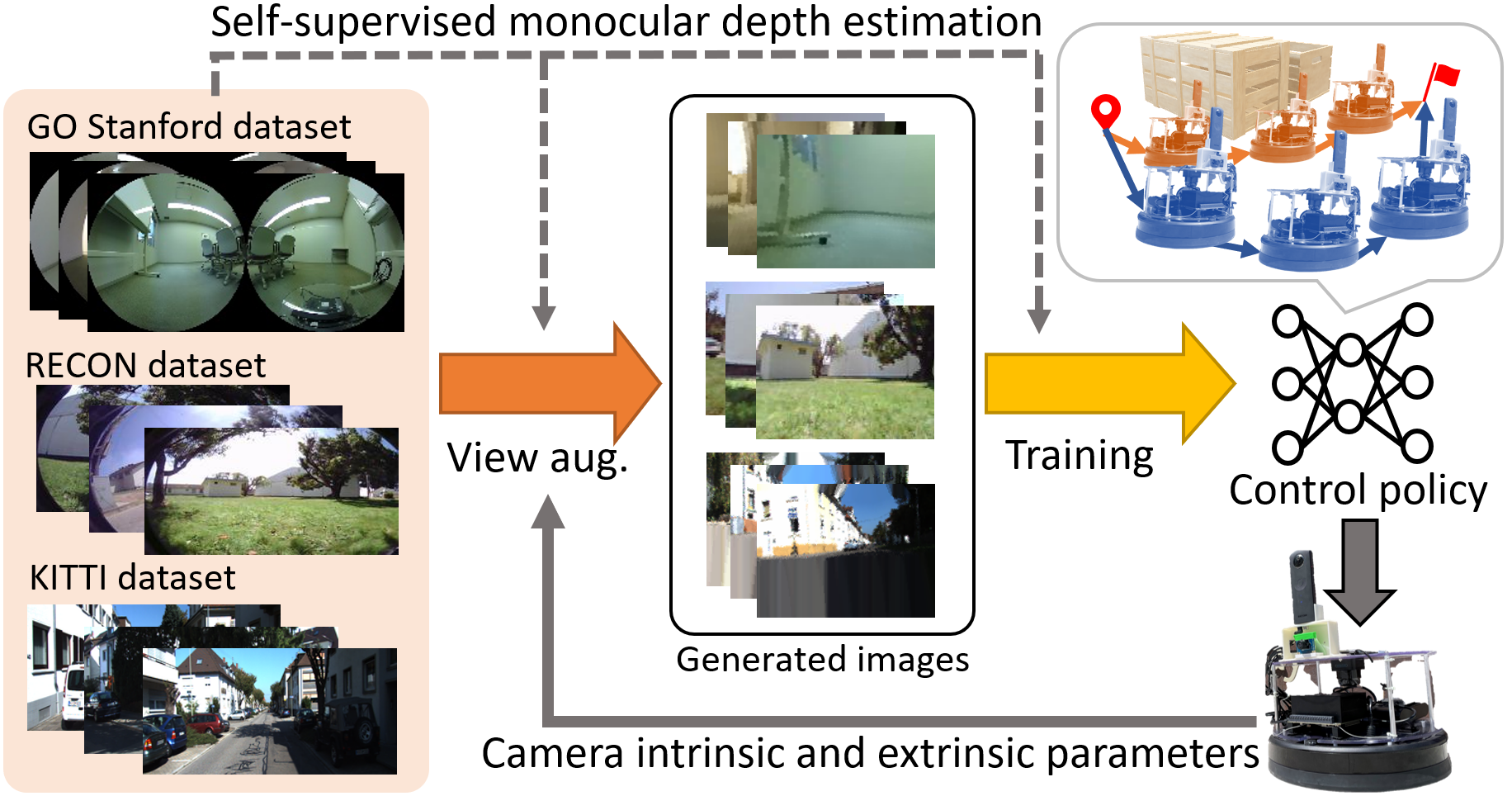}
  \end{center}
      \vspace*{-3mm}
	\caption{\small {\bf Overview of our geometric experience augmentation, \MethodName, for training goal conditioned policy.} Our method synthesizes images with new camera properties in a reconstructed environment via self-supervised monocular depth estimation. The same reconstructed environment is used to train one control policy for a variety of robot parameters.}
  \label{f:overview}
  \vspace*{-5mm}
\end{figure}

Figure~\ref{f:overview} provides an overview of our method. Our challenge is to train a robot conditioned control policy by augmenting experiences from multiple public datasets. Different from \cite{shah2021rapid,hirose2019deep}, our method does not require new dataset collection or online training on our robot. In our method, there are two steps, (1) view augmentation, and (2) geometric-aware policy learning.

To efficiently leverage multiple datasets which are collected by different robots with different cameras, our method generates the images with the specified camera intrinsic and extrinsic parameters via depth estimation in the first step. Note that our method assumes that we can measure the camera intrinsic and extrinsic parameters from our own robot. 
In the second step, we train the policy with a novel geometric-aware objective to avoid collisions between the arbitrary sized robot and its environments. In this training, we use the synthetic images from the first step. 

\subsection{Masked normalization weight in $J_\text{geo}$}
Our geometric-aware objective, $J_\text{geo}$ can penalize the robot positions that collide with the environment. In $J_\text{geo}$ of eq.(\ref{eq:jgeo}), we give the masked normalization weight $g[i, j] = m^g_{k}[i, j] \cdot w^g_{k}[i, j]$. In this section, we explain our implementation of $m^g_{k}[i, j]$ and $w^g_{k}[i, j]$, respectively. 
$m^g_{k}[i, j]$ is the binary value to remove the effect of $(r_s - d_k[i,j])^2$ in the cases where the 3D point $Q_{k}[i,j]$ is outside of the robot's area. 
\begin{eqnarray}
    m^g_{k}[i, j] = \begin{cases}
                1, \hspace{5mm} (\mbox{inside of robot area})\\
                0, \hspace{5mm} (\mbox{outside of robot area})\\
                \end{cases}
    \label{eq:eq_jgeo_mask}
\end{eqnarray}

$w^g_{k}[i, j]$ is the weighting variable to balance the geometric cost according to the sparsity of the 3D points.
\begin{eqnarray}
    w^g_{k}[i, j] &=& f_{dist}(Q_{k}[i-1,j], Q_{k}[i+1,j]) \nonumber \\
    && \times  f_{dist}(Q_{k}[i,j-1], Q_{k}[i,j+1]).
    \label{eq:eq_jgeo}
\end{eqnarray}
Assuming the corresponding 3D point is representative point in the area formed by the adjacent four points, we dictate $w^g_{k}[i, j]$ by its approximate area. Here $f_{dist}(A, B)$ is the function to calculate the distance between A and B. Without $w^g_{k}$, $J_{geo}$ gives larger penalization to obstacles with dense 3D points, and it causes the robot to collide with obstacles with sparse 3D points.  

The other parameters $N_U$ and $N_V$ in eq.(\ref{eq:jgeo}) are the number of pixels on $U$ and $V$ axis. And $L$ is defined as $L = \sum_{k=1}^{N_s}\sum_{j=1}^{N_V}\sum_{i=1}^{N_U}m^{g}_{k}[i,j]$ to calculate the mean of the effective 3D points inside of the robot's area.
%
%
%
\subsection{Process of view synthesis}
Our view synthesis approach generates synthetic images via an estimated point cloud.
As shown in Section \ref{sec:augmenting_geometry}, we project the point clouds onto the image plane. Then, we interpolate the projected images to fill in the pixels without a projected color value. Here, we explain the detail implementation of both the projection and interpolation process.

By projecting the point cloud $Q_{t}$ on to the image plane using intrinsic parameters measured from the target camera, we can get the corresponding pixel position $[i_c, j_c]$ in target image coordinates as follows:
\begin{eqnarray}
    [i_c, j_c] = f_\text{proj}(Q_{t}[i, j]), 
    \label{eq:projx}
\end{eqnarray}
where $f_\text{proj}()$ is the projection function using measured camera intrinsic parameters. Here, $i_c$ and $j_c$ are scalar values (not integer values). Hence, we can have four candidate pixel positions to project the pixel value of $I[i, j]$. To fill in many pixels, we merge four intermediate images $\{ I_{p_i} \}_{k=1\ldots 4}$ in the projection process. 
The four intermediate images can be given as follows:
\begin{eqnarray}
    && \hspace{-5mm} I_{p_1}[\lceil i_{c} \rceil, \lceil j_{c} \rceil] = I[i, j], I_{p_2}[\lceil i_{c} \rceil, \lfloor j_{c} \rfloor] = I[i, j], \nonumber \\
    && \hspace{-5mm} I_{p_3}[\lfloor i_{c} \rfloor, \lceil j_{c} \rceil] = I[i, j], I_{p_4}[\lfloor i_{c} \rfloor, \lfloor j_{c} \rfloor] = I[i, j],  
    \label{eq:projy}
\end{eqnarray}
where $\lceil \cdot \rceil$ and $\lfloor \cdot \rfloor$ indicate ceil and floor functions, respectively. $\{ I_{p_k} \}_{k=1\ldots 4}$ can be obtained by executing eq.(\ref{eq:projy}) for all pixels, starting from the one with the largest depth to consider occlusion. The projected image $I_{p}$ can be calculated by a weighted sum of $\{ I_{p_k} \}_{k=1\ldots 4}$ as follows:
\begin{eqnarray}
    I_{p} = \left( \sum_{k=1}^4 w_{p_k} I_{p_k} \right) /\sum_{k=1}^4 w_{p_k}
    \label{eq:wsum}
\end{eqnarray}
where $\{w_{p_k}\}_{k=1 \ldots 4}$ are weight matrices that give larger weight to the pixels that are closer to $[i_c, j_c]$, 
\begin{eqnarray}
    w_{p_s}= \left( \sum_{k=1}^4 l_k - l_s \right) / \sum_{k=1}^4 l_k
    \label{eq:w_ip}
\end{eqnarray}
where $\{l_k\}_{k=1 \ldots 4}$ are the lengths between $[i_c, j_c]$ and the corresponding pixel position. 
\begin{eqnarray}
    l_1 = \sqrt{(\lceil i_{c} \rceil - i_c)^2 + (\lceil j_{c} \rceil - j_c)^2} \nonumber \\
    l_2 = \sqrt{(\lceil i_{c} \rceil - i_c)^2 + (\lfloor j_{c} \rfloor - j_c)^2} \nonumber \\
    l_3 = \sqrt{(\lfloor i_{c} \rfloor - i_c)^2 + (\lceil j_{c} \rceil - j_c)^2} \nonumber \\
    l_4 = \sqrt{(\lfloor i_{c} \rfloor - i_c)^2 + (\lfloor j_{c} \rfloor - j_c)^2}    
    \label{eq:lk}
\end{eqnarray}

The last process in our view augmentation is interpolation of $I_{p}$. Although we can fill a lot of pixel positions by merging in eq.(\ref{eq:wsum}), there are still a lot of blank pixels in $I_{p}$. Nearest neighbors is one of the simplest methods to fill them. However, it makes the generated images noisy. We fill the blank pixels with a weighted sum of the closest four pixels to generate $I'$, similar to eq.(\ref{eq:wsum}). 

It should be noted that alternative methods are possible in projection and interpolation~\cite{huang2020fast,5662013}. We employ the above process because it is fast on GPU.
\subsection{Navigation system}
To evaluate our proposed method using a real prototype mobile robot in navigation, we construct a navigation system (Algorithm~\ref{alg:nav}) with our trained policy. Our task is to arrive at the position of final goal image $I_{gN_g}$ by using current image $I_c$ from the robot camera and the subgoal images $\{I_{gi}\}_{i=1 \ldots N_g}$ from start to the goal position. 

Before navigation, we teleoperate the robot from the start to the goal position to construct the topological map with $\{I_{gi}\}_{i=1 \ldots N_g}$. We sample $I_{gi}$ every 2 seconds as a new node, and connect an edge with the previously sampled node. After making map, we place the robot around the start position to begin navigation.

Following \cite{hirose2021probabilistic}, there are three modules in our navigation system: 1) localization, 2) the control policy, and 3) the safety modules. The localization module decides the current node number $n_c$ corresponding to most adjacent node and feeds the subgoal image $I_{g(n_c+1)}$ into the control policy module. To decide $n_c$, we estimate the distance $\sqrt{(x^{n_g})^2 + (y^{n_g})^2}$ and the angle $\theta^{n_g}$ between the subgoal position of $I_{gn_g}$ and the current position. Note that we assume that $N_s$-th way point $({x^{n_g}, y^{n_g}, \theta^{n_g}})$ integrated from velocity commands $\{v^{n_g}_i, \omega^{n_g}_i \}_{i=1 \ldots N_s}$ can reach the subgoal position. Here, $x^{n_g}$, $y^{n_g}$ are xy-position and $\theta^{n_g}$ is yaw angle of the robot coordinate. If the estimated subgoal position is close enough to satisfy $\sqrt{(x^{n_g})^2 + (y^{n_g})^2} < d_l$ and $\theta^{n_g} < \theta_l$, we update the current node as $n_c=n_g$. We do this process for $N_t$ subgoal images to allow our system to skip a few subgoals. Trajectories with larger deviations from the original path to avoid collisions occasionally miss some subgoal positions.

In the control policy module, we calculate the velocity commands $\{v^i, \omega^i\}_{i=1 \ldots N_s}$ and the traversability probability $\{t^i \}_{j=1 \ldots N_s}$. Similar to receding horizon control (i.e., model predictive control), we only use $v_1$ and $\omega_1$ after our collision check in the safety module. In safety module, we simply check whether the robot will collide or not by thresholding $p^1$. If our predicted trajectory is untraversable ($p^1 <$ 0.5), we override the linear velocity as 0.0 and only allow the robot to turn at a point. Pivot turning allows the robot to find alternate paths to the subgoal position which are traversable. In our navigation system, we set $N_t$, $d_l$, and $\theta_l$ to 5, 0.4, 0.2, and 0.5, respectively.
\begin{algorithm}
\caption{Our vision-based navigation system.}
\KwData{Subgoal images $\{I_{gi}\}_{i=1 \ldots N_g}$, current image $I_c$, distance threshold $d_l$, angle threshold $\theta_{l}$}
\KwResult{Linear and angular velocity command for a real robot $\{v, \omega \}$}
 $n_c$ = 0 \tcp*{initialization}
 \While{$n_c \neq N_g$}{
  \tcc{Localization module}
  $n_l$ = 1\;
  \While{$n_l \leq N_t$}{
  $n_g = n_c + n_l$ \;
  $\{v^{n_g}_i, \omega^{n_g}_i, p^{n_g}_i \}_{i=1 \ldots N_s} = \pi_\theta(I_c, I_{gn_g}, r_s, v_l)$\;
  $x^{n_g}, y^{n_g}, \theta^{n_g} = f_{ing}(\{v^{n_g}_i, \omega^{n_g}_i \}_{i=1 \ldots N_s})$\;
  \If{$\sqrt{(x^{n_g})^2 + (y^{n_g})^2} < d_l$ and $\theta^{n_g} < \theta_{l}$}{
  $n_c = n_g$ \;
  }
  $n_l = n_l + 1$ \;
  }
  \tcc{Control policy module} 
  $\{v^i, \omega^i, t^i \}_{j=1 \ldots N} = \pi_\theta(I_c, I_{g(n_c + 1)}, r_s, v_l)$ \;
  \tcc{Safety module}   
  \eIf{$p^1 >$ 0.5}{
  $(v, \omega)=(v^1, \omega^1)$\;
  }{
  $(v, \omega)=(0.0, \omega^i)$ \tcp*{pivot turning}
  }
 }
\label{alg:nav}
\end{algorithm}
\vspace*{-3mm}
\subsection{Unseen obstacles in navigation}
\begin{figure}[h]
  \begin{center}
      \includegraphics[width=0.99\hsize]{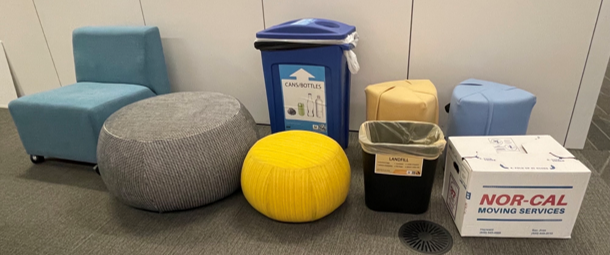}
  \end{center}
      \vspace*{-3mm}
	\caption{\small {\bf Examples of obstacles in navigation.} We randomly pick at least one obstacle and place them after we collect the subgoal images. We place the obstacles on or between the positions of some of the subgoals.}
  \label{f:obstacle}
\end{figure}

We have experiments in six indoor environments and six outdoor environments. In half of the indoor and outdoor environments, we place unseen obstacles in Fig.~\ref{f:obstacle} before starting navigation. In each environment, we have three trials where we vary the initial robot position, kind of obstacles, and the position of these obstacles.
%
\subsection{Visualization of data augmentation}
\begin{figure}[h]
  \begin{center}
      \includegraphics[width=0.99\hsize]{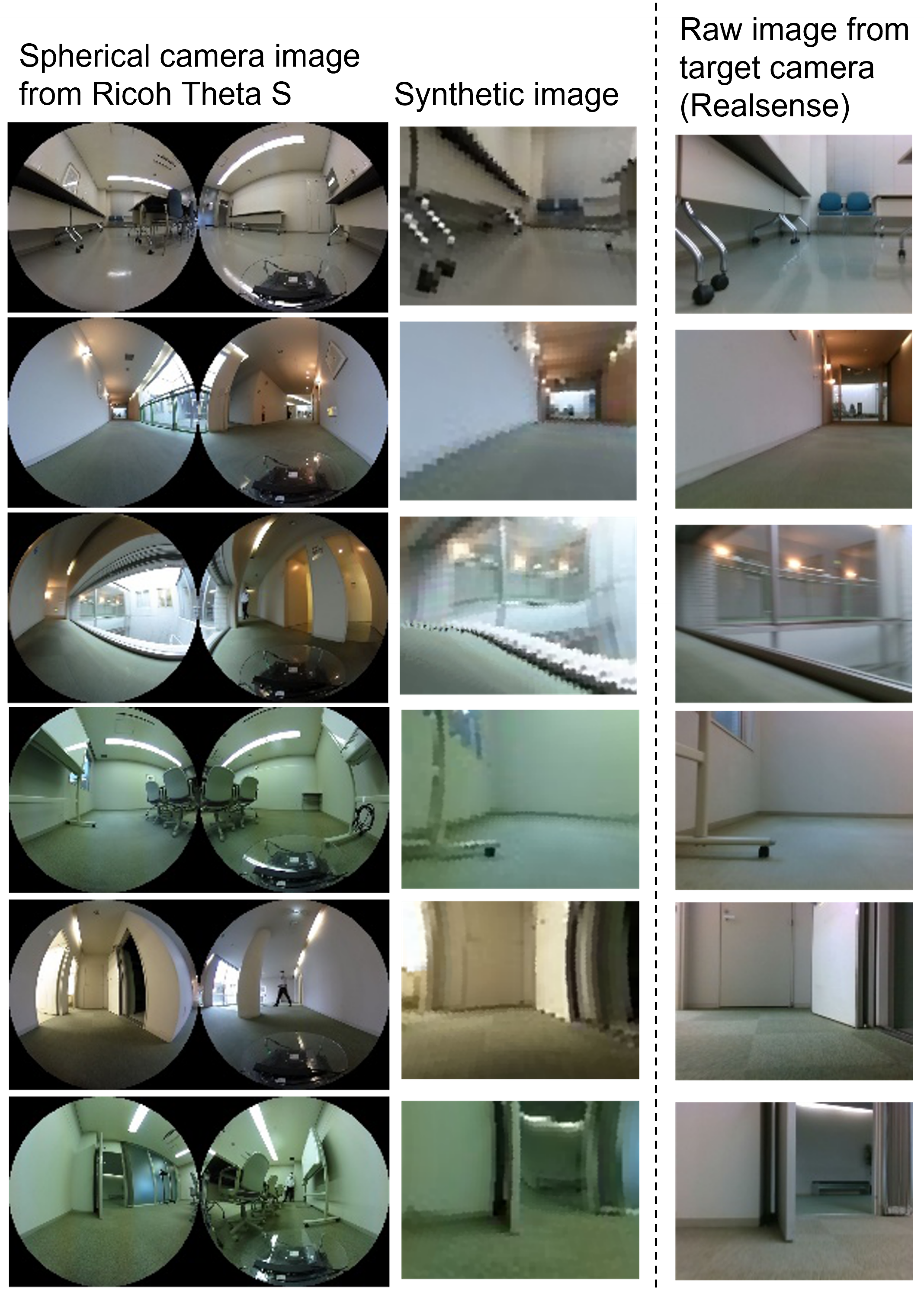}
  \end{center}
      \vspace*{-3mm}
	\caption{\small {\bf Comparison of the synthetic images with the raw images of the target camera, Realsense.} We mount the spherical camera and the Realsense at different positions on our robot platform. We generate synthetic images corresponding to the Realsense from the spherical camera images.}
  \label{f:gen}
\end{figure}

Figure.~\ref{f:gen} shows synthetic images of our view augmentations and raw images of the target camera to visualize our view augmentations. We mounted the spherical camera and the narrow FoV camera at different positions on same robot platform. We generated synthetic images corresponding to the narrow FoV camera from the spherical camera images. Although the generated images are a bit blurry and noisy when comparing them to the raw images, there are enough details to learn the positional relationships between the images.

\end{document}